\documentclass[conference]{IEEEtran}

\IEEEoverridecommandlockouts
\usepackage{cite}
\usepackage{amsmath,amssymb,amsfonts}
\usepackage{algorithmic}
\usepackage{graphicx}
\usepackage{textcomp}
\usepackage{xcolor}
\usepackage{comment}
\usepackage{float}
\usepackage[hyphens]{url}
\usepackage[hidelinks]{hyperref}
\begin{document}

\title{
Enabling Long FFT Convolutions on Memory-Constrained FPGAs via Chunking
}

\author{\IEEEauthorblockN{Peter Wang, Neelesh Gupta, Viktor K. Prasanna}
\IEEEauthorblockA{University of Southern California, Los Angeles, CA \\
\{plwang, neeleshg, prasanna\}@usc.edu}
}
\maketitle

\begin{abstract}
The need for long-context reasoning has led to alternative neural network architectures besides Transformers and self-attention, a popular model being Hyena, which employs causal 1D-convolutions implemented with FFTs. Long convolutions enable efficient global context mixing, but requirements for intermediate results exceed the 2-3 MB Block RAM capacity of FPGAs. We present a chunked FFT convolution approach enabling 450K length sequence by 450K length filter convolutions on an Alveo U200 FPGA with 2.8 MB BRAM through chunking and overlap-add reconstruction.
We find that throughput scales proportionally with chunk size while degrading minimally by 7\% for our longest sequences, demonstrating that careful memory management enables deployment of long-context primitives on edge FPGAs without sacrificing performance. 
\end{abstract}


\section{Introduction}
Advances in sequence modeling have led to convolution-based alternatives to attention mechanisms. Models like Hyena~\cite{hyena} show that long convolutions can match Transformer performance while scaling sub-quadratically with sequence length. 
HyenaDNA~\cite{hyenaDNA} applies this architecture to genomic foundation models, where capturing long-range interactions requires processing sequences up to 1M context length at single nucleotide resolution, necessitating efficient long convolution.

\vspace{-5pt}

\begin{figure}[htbp]
  \centering
  \includegraphics[width=1\columnwidth]{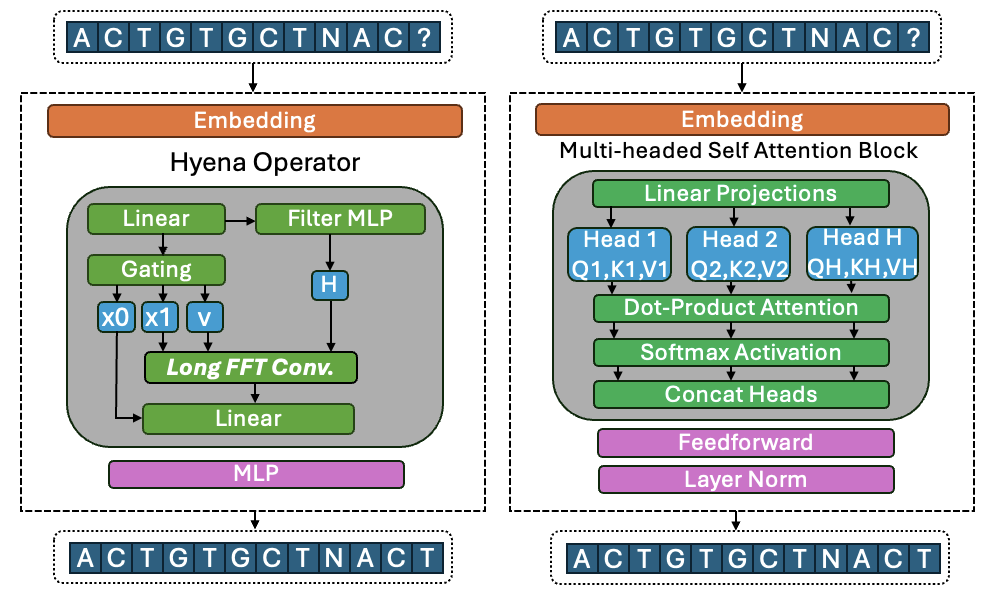}
  \caption{HyenaDNA architecture}
  \label{fig:hyena_image}
\end{figure}

While state-of-the-art implementations like FlashFFTConv~\cite{flashfftconv} have achieved impressive speedups on high-end GPUs, these are impractical for edge deployment cases with constrained power and memory. FPGAs are an attractive alternative through their reconfigurable architecture and superior power efficiency. However, FPGA implementations face the 
\pagebreak
challenge of limited on-chip memory, typically a few MB of Block RAM (BRAM) compared to gigabytes of GPU VRAM. 

This work addresses the FPGA memory constraint through \textit{chunked FFT convolution}: decomposing large convolutions into chunks that fit within BRAM constraints, processing each chunk independently, and reconstructing the final result via overlap-add. Our contributions are as follows:

\begin{itemize}
\item We implement a chunked FFT convolution algorithm enabling up to 450K sequence length by 450K filter length 1D-convolutions on FPGAs with 2.8 MB BRAM.
\item We analyze end-to-end performance, finding that FFT convolution increasingly dominates execution time, from 27\% for our shortest sequence to 35\% for our longest. 
\item We evaluate performance across chunk sizes and sequence lengths, finding that throughput depends on chunk size and that it degrades by about 7\% as sequence length increases from 32K to 450K.
\end{itemize}

\section{Background}
\subsection{FFT-Based Convolution Latency in Hyena} 
We tested FFT-Convolution and the Hyena Block on GPU. Latency for convolution compared to the whole block increases from about 27\% for the smallest sequence length to 35\% for the longest, demonstrating that FFT-Convolution dominates in longer contexts, making it the main challenge.  

While existing libraries like cuFFT have been previously shown to complete an 8M point FFT in about 0.2 seconds on an A100 (300W TDP, 1.4 GHz clock rate), we aim to eventually enable up to 1M context FFT-Convolution on an FPGA U200 (200W TDP, 300MHz clock rate). 

\subsection{FPGA Memory}
Block RAM (BRAM) provides uniform low-latency random access, making it ideal for portions (such as FFT butterfly) that exhibit data-dependent, non-sequential access patterns with poor spatial locality. The constraint is the limited capacity; the Xilinx Alveo U200 which we use offers 2.8 MB BRAM.

\subsection{Problem Definition}
For input $x[n]$ of length $N_x$ and filter $h[n]$ of length $N_h$, direct convolution requires $O(N_x \cdot N_h)$ operations, which is prohibitive for long sequences where the filter length matches input length, as in Hyena. 
Our objective is computing the convolution $y[n]$ of length $N_y=N_x+N_h-1$ on FPGA using the FFT-based method instead, which runs in $O(N \log N)$. 

\begin{equation}
y[n] = \mathcal{F}^{-1}\{\mathcal{F}\{x[n]\} \cdot \mathcal{F}\{h[n]\}\}
\end{equation}

\section{Proposed Methodology}
We decompose the problem into (1) a radix-2 FFT kernel optimized for on-chip memory constraints and (2) a host-side layer managing data-transfer and overlap-add reconstruction.
\subsection{FFT Kernel Design}

We implement the Cooley-Tukey radix-2 decimation-in-time FFT algorithm on an Alveo U200 FPGA with Vitis HLS. The kernel processes chunks up to 8,192 points, determined experimentally. We adopt this strategy to enable longer-context sequence modeling primitives at lower energy cost.

\vspace{-8pt}

\begin{figure}[htbp]
  \centering
  \includegraphics[width=0.9\columnwidth]{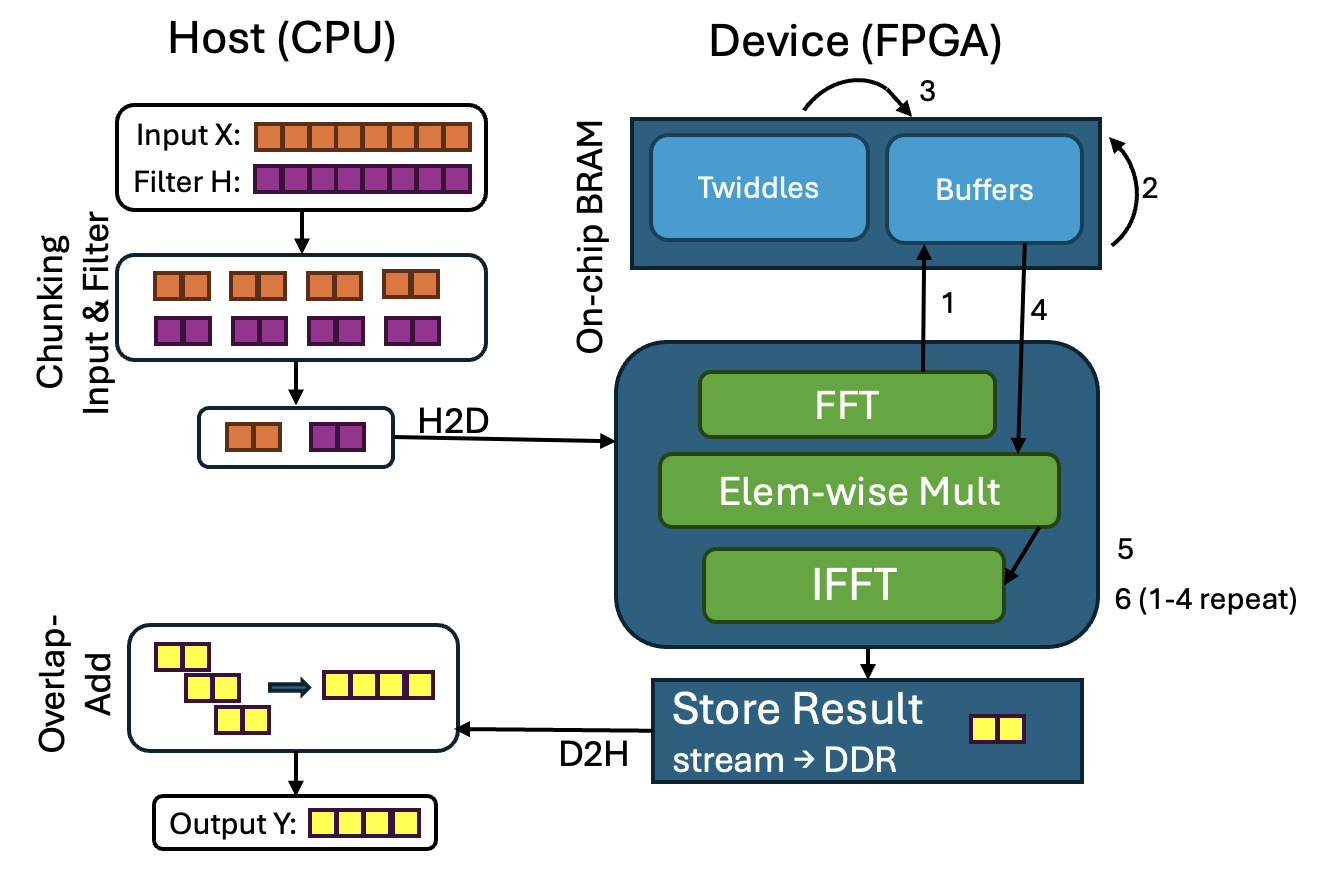}
  \caption{Our CPU-host and FPGA-kernel implementation}
  \label{fig:architecture_image}
\end{figure}
\vspace{-6pt}

\subsection{Host Side Chunked Convolution Strategy}

We partition the input sequence $x[N_x]$ and filter $h[N_h]$ into:
\begin{align}
x[n] &\rightarrow \{x_0[n], x_1[n], \ldots, x_{M_x-1}[n]\} \\
h[n] &\rightarrow \{h_0[n], h_1[n], \ldots, h_{M_h-1}[n]\}
\end{align}
with $M_x = \lceil N_x / C \rceil$ and $M_h = \lceil N_h / C \rceil$ for chunk size $C$.

Each chunk pair is multiplied in the frequency domain, and then 
the final output is reconstructed via overlap-add:
\begin{equation}
y[n] = \sum_{i=0}^{M_x-1} \sum_{j=0}^{M_h-1} y_{i,j}[n - iC - jC]
\end{equation}

\section{Preliminary Results}
We test on sequences from the hg38 human 
reference genome with filters of equal length generated in each case. Sequence length of our tests follows those in HyenaDNA. 

\subsection{Performance Results}

Table I shows execution time across chunk sizes and sequence lengths. 

\begin{table}[h]
\centering
\caption{Execution Time (seconds) for Different Chunk Sizes}
\label{tab:performance}
\begin{tabular}{|c|c|c|c|}
\hline
\textbf{Dataset Size} & \textbf{8K Chunk} & \textbf{4K Chunk} & \textbf{2k Chunk} \\
\hline
450K by 450K & 47 & 188 & 750 \\
160K by 160K & 6 & 27 & 96 \\
32K by 32K & 1 & 1 & 3 \\
\hline
\end{tabular}
\end{table}

Table II shows throughput increasing with chunk size. 
In a given chunk size, throughput is highest among smaller inputs, with an average degradation of about 7\% from smallest to largest inputs. Bandwidth for all tests was about 0.8 GB/s. 

\begin{table}[h]
\centering
\caption{Throughput (MFLOPS) by chunk size and dataset size}
\label{tab:performance}
\begin{tabular}{|c|c|c|c|}
\hline
\textbf{Dataset Size} & \textbf{8K Chunk} & \textbf{4K Chunk} & \textbf{2k Chunk} \\
\hline
450K by 450K & 104 & 46 & 22 \\
160K by 160K & 109 & 48 & 24 \\
32K by 32K & 110 & 50 & 25 \\
\hline
\end{tabular}
\end{table}

\subsection{Compute vs. Transfer Analysis}

Table~\ref{tab:breakdown} decomposes execution time into kernel lauch and execution, data transfer, and CPU processing.

\vspace{-8 pt}

\begin{table}[H]
\centering
\caption{Time Breakdown: Compute / H2D+D2H Transfer / CPU (\%)}
\label{tab:breakdown}
\begin{tabular}{|c|c|c|c|}
\hline
\textbf{Dataset Size} & \textbf{8K Chunk} & \textbf{4K Chunk} & \textbf{2K Chunk}\\
\hline
450K & 97.9 / 1.4 / 0.6 & 98.2 / 1.4 / 0.4  & 98.4 / 1.3 / 0.2\\
160K & 97.9 / 1.4 / 0.6 & 98.2 / 1.4 / 0.3 & 98.2 / 1.5 / 0.2\\
32K & 98.0 / 0.8 / 0.6 & 98.3 / 1.3 / 0.3 & 98.3 / 1.4 / 0.2\\
\hline
\end{tabular}
\end{table}

\subsection{Performance Comparison}
Recent FPGA-based implementations have achieved high throughput for standalone FFT operations. Li et al.~\cite{li_hls_fft} 
achieved 94 MSamples/s on the Virtex4 with their configurable architecture for arbitrary-length FFTs, while Bai et al.~\cite{dsp-opu} achieved up to 400 MSamples/s for 1024-point FFT operations on the Alveo U200 at 280 MHz operating frequency compared to our maximum of about 10 MSamples/s. While we are working with long convolutions with more overhead, these numbers and our results showing compute dominating our workflow indicate room for optimization in our approach. 

\section{Conclusion}
We presented an approach that enables long FFT convolutions on  FPGAs, showing that they serve as viable accelerators for memory-intensive deep learning operations. Employing chunked convolution, we processed up to 450K length sequence by 450K length filter convolutions on FPGA hardware with limited BRAM. Results show compute dominating performance and throughput scaling proportionally with chunk size, while maintaining performance across workloads. 

Our next step is designing a sparse causal 1D-convolution operator to reduce compute and memory requirements, which has not been done to our best knowledge. Other optimizations include integrating URAM alongside BRAM and deploying parallel FFT kernels to process chunk pairs simultaneously. 

\section*{Acknowledgement}
This work is supported by the NSF under grant numbers CNS-2009057 and CSSI-2311870. 

\bibliographystyle{IEEEtran}
\bibliography{references}

\end{document}